\pdfoutput=1
\documentclass[10pt,twocolumn,letterpaper]{article}

\usepackage[pagenumbers]{cvpr} 

\usepackage{graphicx}
\usepackage{amsmath}
\usepackage{amssymb}
\usepackage{booktabs}
\usepackage{xcolor}
\newcommand{\bp}[1]{\textcolor{black}{#1}}

\makeatletter
\renewcommand{\maketag@@@}[1]{\hbox{\m@th\normalsize\normalfont#1}}%
\makeatother
\usepackage{algorithm}  
\usepackage{algorithmic} 

\usepackage[pagebackref,breaklinks,colorlinks]{hyperref}
\usepackage{multirow}

\usepackage[capitalize]{cleveref}
\crefname{section}{Sec.}{Secs.}
\Crefname{section}{Section}{Sections}
\Crefname{table}{Table}{Tables}
\crefname{table}{Tab.}{Tabs.}

\def\cvprPaperID{8178} 
\def\confName{CVPR}
\def\confYear{2022}

\begin{document}

\title{$\beta$-DARTS: Beta-Decay Regularization for Differentiable Architecture Search}

\author{Peng Ye\textsuperscript{1}\thanks{part of this work was done when Ye Peng was tele-interned at Baidu.}, Baopu Li\textsuperscript{2}, Yikang Li\textsuperscript{3}, Tao Chen\textsuperscript{1}\thanks{ Corresponding author}~, Jiayuan Fan\textsuperscript{1}, Wanli Ouyang\textsuperscript{4}\\
\textsuperscript{1}Fudan University, \textsuperscript{2}BAIDU USA LLC,  \\
\textsuperscript{3}Shanghai AI Laboratory, \textsuperscript{4}The University of Sydney\\
{\tt\small yepeng20@fudan.edu.cn}
}

\maketitle
\begin{abstract}
Neural Architecture Search~(NAS) has attracted increasingly more attention in recent years because of its capability to design deep neural network automatically. 
Among them, differential NAS approaches such as DARTS, have gained popularity for the search efficiency. 
However, they suffer from two main issues, the weak robustness to the performance collapse and the poor generalization ability of the searched architectures. 
To solve these two problems, a simple-but-efficient regularization method, termed as Beta-Decay, is proposed to regularize the DARTS-based NAS searching process. Specifically, Beta-Decay regularization can impose constraints to keep the value and variance of activated architecture parameters from too large.
Furthermore, we provide in-depth theoretical analysis on how it works and why it works. 
Experimental results on NAS-Bench-201 show that our proposed method can help to stabilize the searching process and makes the searched network more transferable across different datasets. 
In addition, our search scheme shows an outstanding property of being less dependent on training time and data.
Comprehensive experiments on a variety of search spaces and datasets validate the effectiveness of the proposed method.
\end{abstract}

\section{Introduction} \label{sec:intro}
\begin{figure}[t] 
	\centering
	\includegraphics[width=3.2in]{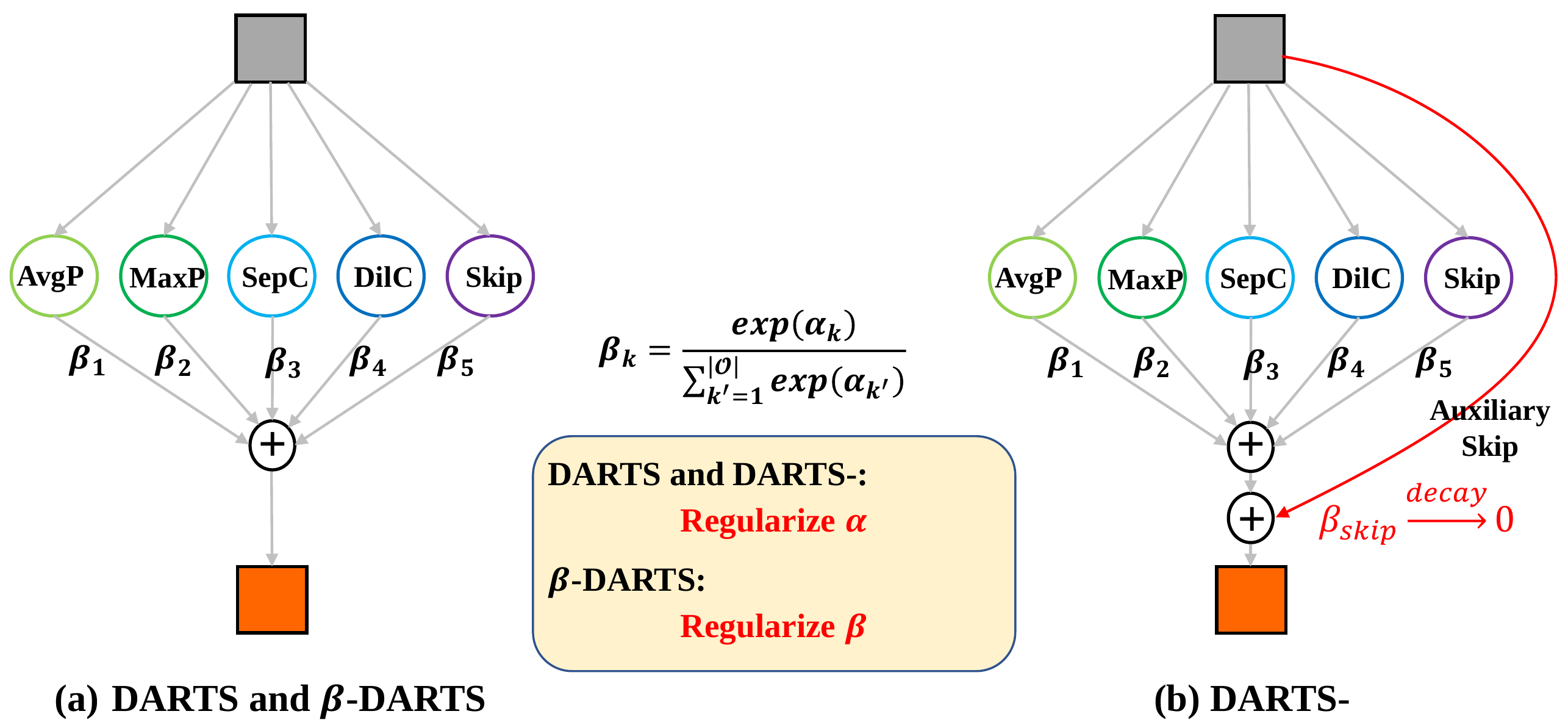}
	\vspace{-12pt}
	\caption{ Schematic illustration about (a) DARTS~\cite{darts} and our proposed $\beta$-DARTS, (b) DARTS-~\cite{darts-}.
	DARTS- adds an auxiliary skip connection with a decay rate $\beta_{skip}$
	to alleviate the performance collapse problem. $\beta$-DARTS introduces the Beta-Decay regularization to improve both the robustness of the searching process and the generalization ability of the searched architecture.}
	\label{fig:1}
	\vspace{-12pt}	
\end{figure}

Neural architecture search (NAS) has attracted lots of interests for its potential to automatize the process of architecture design. Previous reinforcement learning~\cite{nasnet, mnasnet} and evolutionary algorithm~\cite{amoebanet} based methods usually incur massive computation overheads, which hinder their practical applications. To reduce the search cost, a variety of approaches are proposed, including performance estimation~\cite{klein2016learning}, network morphisms~\cite{cai2018path} and one-shot architecture search~\cite{spos,darts}. In particular, one-shot methods resort to weight sharing technique, which only needs to train a supernet covering all candidate sub-networks once. Based on this weight sharing strategy, differentiable architecture search~\cite{darts} (namely DARTS, as shown in Fig.~\ref{fig:1}) relaxes the discrete operation selection problem to learn differentiable architecture parameters, which further improves the search efficiency by alternately optimizing supernet weights and architecture parameters.

Although differentiable method has the advantages of simplicity and computational efficiency, its robustness and architecture generalization challenges still needs to be fully resolved. 
Firstly, lots of studies have shown that DARTS frequently suffers from performance collapse, that is the searched architecture tends to accumulate parameter-free operations especially for skip connection, leading to the performance degradation~\cite{darts-,rdarts}. To handle this robustness challenge, lots of instructive works are proposed: directly restricting the number of skip connections~\cite{pdarts,darts+}; exploiting or regularizing relevant indicators such as the norm of Hessian regarding the architecture parameters~\cite{rdarts,sdarts}; changing the searching and/or discretization process~\cite{fairdarts,dots,darts-}; implicitly regularizing the learned architecture parameters~\cite{rdarts}. However, the explicit regularization of architecture parameters optimization receives little attention, as previous works (including above methods) adopt L2 or weight decay regularization by default on learnable architecture parameters (i.e., $\alpha$), without exploring solution along this direction. 
Secondly, several works have pointed out that the optimal architecture obtained on the specific dataset cannot guarantee its good performance on another dataset~\cite{adaptNAS,mixsearch}, namely the architecture generalization challenge. To improve the generalization of searched model, AdaptNAS~\cite{adaptNAS} explicitly minimizes the generalization gap of architectures between domains via the idea of cross domain,
MixSearch~\cite{mixsearch} searches a generalizable architecture by mixing multiple datasets of different domains and tasks. 
However, both methods solve this issue by leveraging larger datasets, while how to use a single dataset to learn a generalized architecture remains challenging. 

This paper is dedicated to simultaneously solve the above-mentioned two challenges in an efficient way. 
Inspired by the widely-used L2~\cite{cortes2012l2} or weight decay regularization~\cite{wd} approaches, we intend to design a customized regularization for DARTS-based methods, which can explicitly regularize the optimizing process of architecture parameters. However, different from the regularization on the learnable architecture parameter set, $\alpha$ (before the nonlinear activation of softmax), commonly used in standard DARTS and its subsequent variants, we propose a novel and generic Beta-Decay regularization, imposing regularization on the activated architecture parameters $\beta$ (after softmax), where $\beta_k=\frac{\exp \left(\alpha_{k}\right)}{\sum_{k^{\prime}=1}^{\left | \mathcal{O}\right |} \exp \left(\alpha_{k^{\prime}}\right)}$.
On one hand, the proposed Beta-Decay regularization is very simple to implement, achieved with only additional one line of PyTorch code in DARTS~(Alg~\ref{pytorch_implementation}). On the other hand, this simple implementation is grounded by in-depth theoretical support. We provide theoretical analysis to show that, Beta-decay regularization not only mitigates unfair competition advantage among operations and solve the domination problem of parameter-free operations, but also minimizes the Lipschitz constraint defined by architecture parameters and make sure the generalization ability of searched architecture. In addition, we mathematically and experimentally demonstrate that, commonly-used L2 or weight decay regularization on $\alpha$ may not be effective or even counterproductive for improving robustness and generalization of DARTS. 
\vspace{-6pt}
\begin{algorithm}[htb]
\label{alg:beta-decay code}   
\begin{algorithmic}[1] 
\STATE $\mathcal{L}_{Beta}=$  torch.mean(torch.logsumexp( \\
\qquad \quad self.model.\_arch\_parameters, dim=-1))
\STATE loss $=$ self.\_val\_loss(self.model, input\_valid, \\ 
\qquad \quad target\_valid)+$\lambda\mathcal{L}_{Beta} $
\end{algorithmic}  
\caption{PyTorch Implementation in DARTS}
\label{pytorch_implementation}
\end{algorithm} 
\vspace{-12pt}

DARTS with Beta-Decay regularization ($\beta$-DARTS) is illustrated in Fig.~\ref{fig:1}. Extensive experiments on various search spaces (i.e. NAS-Bench-201, DARTS, NAS-Bench-1Shot1) and datasets (i.e. CIFAR-10, CIFAR-100, ImageNet) verify the effectiveness of our method. Besides, our search scheme shows the following outstanding properties:
 \begin{itemize}
\vspace{-8pt}
\item The search trajectories on NAS-Bench-201 and NAS-Bench-1Shot1 show that, the found architecture has continuously rising performance, and the search process can reach its optimal point at an early epoch.
\vspace{-8pt}
\item We only need to search once on the proxy dataset (i.e., CIFAR-10), but the searched architecture can obtain promising performance on various datasets (i.e., CIFAR-10, CIFAR-100 and ImageNet).
\end{itemize}

\section{Related Works} \label{sec:related work}
\subsection{Robustness of DARTS}
As DARTS is known to chronically suffer from the performance collapse issue caused by the domination of parameter-free operators, lots of works have dedicated to resolving it. P-DARTS~\cite{pdarts} and DARTS+~\cite{darts+} directly limit the number of skip connections. Such handcrafted rules are somewhat suspicious and may mistakenly reject good networks. R-DARTS~\cite{rdarts} finds that the Hessian eigenvalues can be regarded as an indicator for the collapse, and employs stronger regularization or augmentation on the training of supernet weights to reduce this value. Then SDARTS~\cite{sdarts} implicitly regularizes this indicator by adding perturbations to architecture parameters via random smoothing or adversarial attack. Both methods are indirect solutions and rely heavily on the quality of the indicator. FairDARTS~\cite{fairdarts} avoids operation competition by weighting each operation via independent sigmoid function, which will be pushed to zero or one by an MSE loss. DropNAS~\cite{dropnas} proposes a grouped operation dropout for the co-adaption problem and matthew effect. DOTS~\cite{dots} further uses the group operation search scheme to decouple the operation and topology search. DARTS-~\cite{darts-} factors out the optimization advantage of skip connection by adding an auxiliary one. However, these methods circumvent the domination effect of parameter-free operations by changing the searching and/or discretization process or adding extra parameters. Different from these works, we explore a more generic solution by explicitly regularizing the architecture parameters optimization, making original DARTS great again.
\subsection{Generalization of DARTS}
Improving the generalization ability of deep model has always been the focus of deep learning research. Recent works provide guarantee on model generalization by minimizing loss value and loss sharpness simultaneously~\cite{foret2020sharpness}. However, the model generalization is not only related to the network weights, but also determined by its architecture. To this end, several methods attempt to improve the generalization of architectures in the field of NAS. AdaptNAS~\cite{adaptNAS} incorporates the idea of domain adaptation into the search process of DARTS, which can minimize the generalization gap of neural architectures between domains. 
MixSearch~\cite{mixsearch} uses a composited multi-domain multi-task dataset to search a generalizable architecture in a differentiable manner. On one hand, both above methods are built on the assumption of having multiple datasets, while our method is not built on multiple datasets. 
On the other hand, our focus is on regularizing architecture parameters, which is not investigated in AdaptNAS and MixSearch.

\section{Proposed method} \label{sec:method}
\subsection{Formulation of DARTS}
Following \cite{zoph2018learning}, DARTS searches the structure of normal cell and reduction cell to stack the full network. Typically, a cell is defined as a directed acyclic graph (DAG) with \textit{N} nodes, where each node denotes a latent representation and the information between every two nodes is transformed by an edge. Each edge $(i, j)$ contains several candidate operations, and DARTS applies continuous relaxation via the learnable architecture parameters set $\alpha$ to mix the outputs of different operations, converting the discrete operation selection into a differentiable parameter optimization problem,
\begin{equation}\label{eqn1} 
\scriptsize
  \begin{split}
    &\bar{O}^{(i, j)}(x)=\sum_{k=1}^{\left | \mathcal{O}\right |} \beta_{k}^{(i, j)} O_k(x),\quad \beta_{k}^{(i, j)}=\frac{\exp \left(\alpha_{k}^{(i, j)}\right)}{\sum_{k^{\prime}=1}^{\left | \mathcal{O}\right |} \exp \left(\alpha_{k^{\prime}}^{(i, j)}\right)}
  \end{split}
  \vspace{-6pt}
\end{equation} 
where $x$ and $\bar{O}$ are the input and mixed output of an edge, $\mathcal{O}$ is the candidate operation set, and $\beta$ denotes the softmax-activated architecture parameter set. In this way, we can perform architecture search in a differentiable manner by solving following bi-level optimization objective,
\begin{equation} \label{eqn2}
  \begin{split}
    &\min_{\alpha}\mathcal{L}_{val}\left(w^{*}(\alpha), \alpha\right) \\
    &\text { s.t. } w^{*}(\alpha)=\arg \min _{w} \mathcal{L}_{train}(w, \alpha)
  \end{split}
  \vspace{-6pt}
\end{equation}
In practice, architecture parameters $\alpha$ and network weights $w$ are alternately updated on the validation and training datasets via gradient descent, and $w^{*}$ is approximated by one-step forward or current $w$~\cite{darts}.

\subsection{Commonly-used Regularization}
In this paper, we intend to improve the robustness and architecture generalization of DARTS by explicitly regularizing the optimizing process of architecture parameters. Thus, we begin with the default settings of previous methods, namely L2 or weight decay regularization on architecture parameters, $\alpha$. For convenience of analysing, we consider the single-step update of the architecture parameters,
\begin{equation} \label{eqn3}
  \begin{split}
    \alpha^{t+1}_k \leftarrow \alpha^{t}_k-\eta_{\alpha} \cdot \nabla_{\alpha_k} \mathcal{L}_{val}
  \end{split}
  \vspace{-6pt}
\end{equation}
where $\eta_{\alpha}$ and $\mathcal{L}_{val}$ are the learning rate of architecture parameters and the corresponding loss respectively. For multi-step updates, it can be transformed into a single-step update problem through step-wise recursive analysis.

In standard DARTS and its subsequent variants, Adam optimizer with L2 regularization is commonly used for architecture parameters optimization. However, for adaptive gradient algorithms, the gradients of L2 regularization are normalized ($\mathcal{N}$) by their summed magnitudes, thus the penalty for each element is relatively even, which partly offsets the effect of L2 regularization~\cite{dwd}, defined as
\begin{equation} \label{eqn4}
  \begin{split}
    \bar{\alpha}^{t+1}_k \leftarrow \alpha^{t}_k-\eta_{\alpha} \cdot \nabla_{\alpha_k} \mathcal{L}_{val}-\eta_{\alpha} \lambda \mathcal{N}\left(\alpha^{t}_k\right)
  \end{split}
  \vspace{-6pt}
\end{equation}

Considering that L2 regularization may not be effective in adaptive gradient algorithms and is not identical to weight decay regularization~\cite{dwd}, without loss of generality, we also include architecture parameters optimization with weight decay regularization~\cite{hanson1988comparing} for comparison, defined as
\begin{equation} \label{eqn6}
  \begin{split}
    \bar{\alpha}^{t+1}_k \leftarrow \alpha^{t}_k-\eta_{\alpha} \cdot \nabla_{\alpha_k} \mathcal{L}_{val}-\eta_{\alpha} \lambda \alpha^{t}_k
  \end{split}
  \vspace{-6pt}
\end{equation}

\begin{figure*}[t] 
	\centering
	\includegraphics[width=1.0\linewidth]{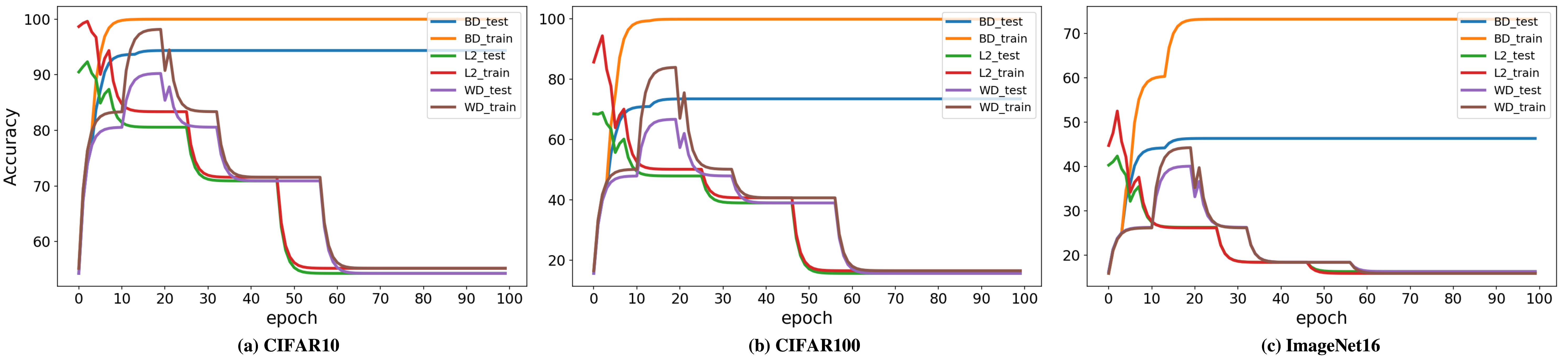}
	\vspace{-12pt}
	\caption{Accuracy of different datasets
	of DARTS with L2, Weight Decay (WD) and Beta-Decay (BD) regularization on NAS-Bench-201 benchmark. The curve is smoothed with a coefficient of 0.5. Note that we only search once on CIFAR-10 dataset.}
	\label{fig:3}
	\vspace{-10pt}	
\end{figure*}

\subsection{Beta-Decay Regularization}
Since the searching and discretization process of DARTS actually utilize softmax-activated architecture parameter set, $\beta$, to represent the importance of each operator, we shall pay more attention to the explicit regularization on $\beta$. 
As shown in Subsection~\ref{sec:Theo}, Beta regularization has the ability to improve the robustness and architecture generalization of DARTS, which further denotes its significance. Although important, Beta regularization is typically ignored by previous works. This paper is devoted to filling this gap. Similar to the idea of most regularization methods, the core purpose of Beta regularization is to constrain the value of Beta from changing too much, formulated as
\begin{equation} \label{eqn8}
\small
  \begin{split}
    \bar{\beta}_{k}^{t+1}=\theta_{k}^{t+1}\left(\alpha_k^t \right)\beta_{k}^{t+1}
  \end{split}
  \vspace{-6pt}
\end{equation}

For simplicity, we use a $\theta$ function with $\alpha$ as the independent variable to express the total influence of Beta regularization here. To realize above Beta regularization similar to weight decay through $\alpha$, we firstly study the influence of $\alpha$ regularization on $\beta$. Recalling Eq.~(\ref{eqn4}) and Eq.~(\ref{eqn6}), we can conclude a unified formula as: 
\begin{equation} \label{eqn19}
  \begin{split}
    \bar{\alpha}_{k}^{t+1} \leftarrow \alpha_{k}^{t}-\eta_{\alpha} \nabla_{\alpha_{k}} \mathcal{L}_{val}-\eta_{\alpha} \lambda F\left(\alpha_{k}^{t}\right)
  \end{split}
  \vspace{-6pt}
\end{equation}

Further, we substitute Eq.~(\ref{eqn19}) and Eq.~(\ref{eqn3}) into Eq.~(\ref{eqn1}) to get $\bar{\beta}_{k}^{t+1}$ and $\beta_{k}^{t+1}$, and then divide the former by the latter.
\begin{equation} \label{eqn9}
\footnotesize
  \begin{split}
    \frac{\bar{\beta}_{k}^{t+1}}{\beta_{k}^{t+1}}=\frac{\sum_{k^{\prime}=1}^{\left | \mathcal{O}\right |} \exp \left(\alpha_{k^{\prime}}^{t+1}\right)}{\sum_{k^{\prime}=1}^{\left | \mathcal{O}\right |} \left[\exp \left(F(\alpha_{k}^{t})-F(\alpha_{k^{\prime}}^{t})\right)\right]^{\lambda \eta_{\alpha}} \exp \left(\alpha_{k^{\prime}}^{t+1}\right)}
  \end{split}
  \vspace{-6pt}
\end{equation}

As we can see in Eq.~(\ref{eqn9}), the mapping function $F$ determines the influence of $\alpha$ on $\beta$. Thus, all we need is to look for a suitable mapping function, $F$. Intuitively, a satisfactory $F$ should meet the following two points: (1) $F$ is not affected by the amplitude of $\alpha$ (to avoid invalid regularization and optimization difficulties). 
(2) $F$ can reflect the relative amplitude of $\alpha$ (to impose more penalty on larger amplitude). 
To satisfy the two requirements, we adopt the softmax to normalize $\alpha$, 
\begin{equation} \label{eqn11}
  \begin{split}
    F\left(\alpha_{k}\right)=\frac{\exp \left(\alpha_{k}\right)}{\sum_{k^{\prime}=1}^{\left | \mathcal{O}\right |} \exp \left(\alpha_{k^{\prime}}\right)}
  \end{split}
  \vspace{-6pt}
\end{equation}
Then we introduce our proposed Beta-Decay regularization loss, whose gradients with respective to $\alpha$ equals to $F\left(\alpha\right)$, 
\begin{equation} \label{eqn10}
\small
  \begin{split}
    \mathcal{L}_{Beta} = \log \left(\sum_{k=1}^{\left | \mathcal{O}\right |}  e^{\alpha_k}\right) = \operatorname{smoothmax} \left(\left\{\alpha_{k}\right\}\right) 
  \end{split}
  \vspace{-6pt}
\end{equation}

After that, substituting Eq.~(\ref{eqn11}) into Eq.~(\ref{eqn9}), we can obtain following equation, which further accounts for the effect of Beta-Decay regularization,
\begin{equation} \label{eqn12}
\footnotesize
  \begin{split}
    \theta_{k}^{t+1}\left(\alpha_k^t \right)=\frac{\sum_{k^{\prime}=1}^{\left | \mathcal{O}\right |} \exp \left(\alpha_{k^{\prime}}^{t+1}\right)}{\sum_{k^{\prime}=1}^{\left | \mathcal{O}\right |} \left[\exp \left(\frac{\exp(\alpha_{k}^{t})-\exp(\alpha_{k^{\prime}}^{t})}{\sum_{k^{\prime\prime}=1}^{\left | \mathcal{O}\right |} \exp \left(\alpha_{k^{\prime\prime}}^{t}\right)}\right)\right]^{\lambda \eta_{\alpha}} \exp \left(\alpha_{k^{\prime}}^{t+1}\right)}
  \end{split}
  \vspace{-6pt}
\end{equation}
Observing the above formula, we can get following conclusions: (1) When $\alpha$ is the largest, $\theta$ is the smallest and less than 1; when $\alpha$ is the smallest, $\theta$ is the largest and greater than 1; and when $\alpha$ is equal, $\theta=1$. (2) In current iteration, $\theta$ decreases as $\alpha$ increases. (3) $\theta$ is smaller when $\alpha$ is larger, and $\theta$ is larger when $\alpha$ is smaller. As a result, the variance of $\beta$ is constrained to be smaller, and the value of $\beta$ is constrained to be closer to its mean, achieving the effect similar to weight decay, thus called Beta-Decay regularization.

\subsection{Theoretical Analysis} \label{sec:Theo}
  \begin{figure}[t] 
	\centering
	\includegraphics[width=3.0in]{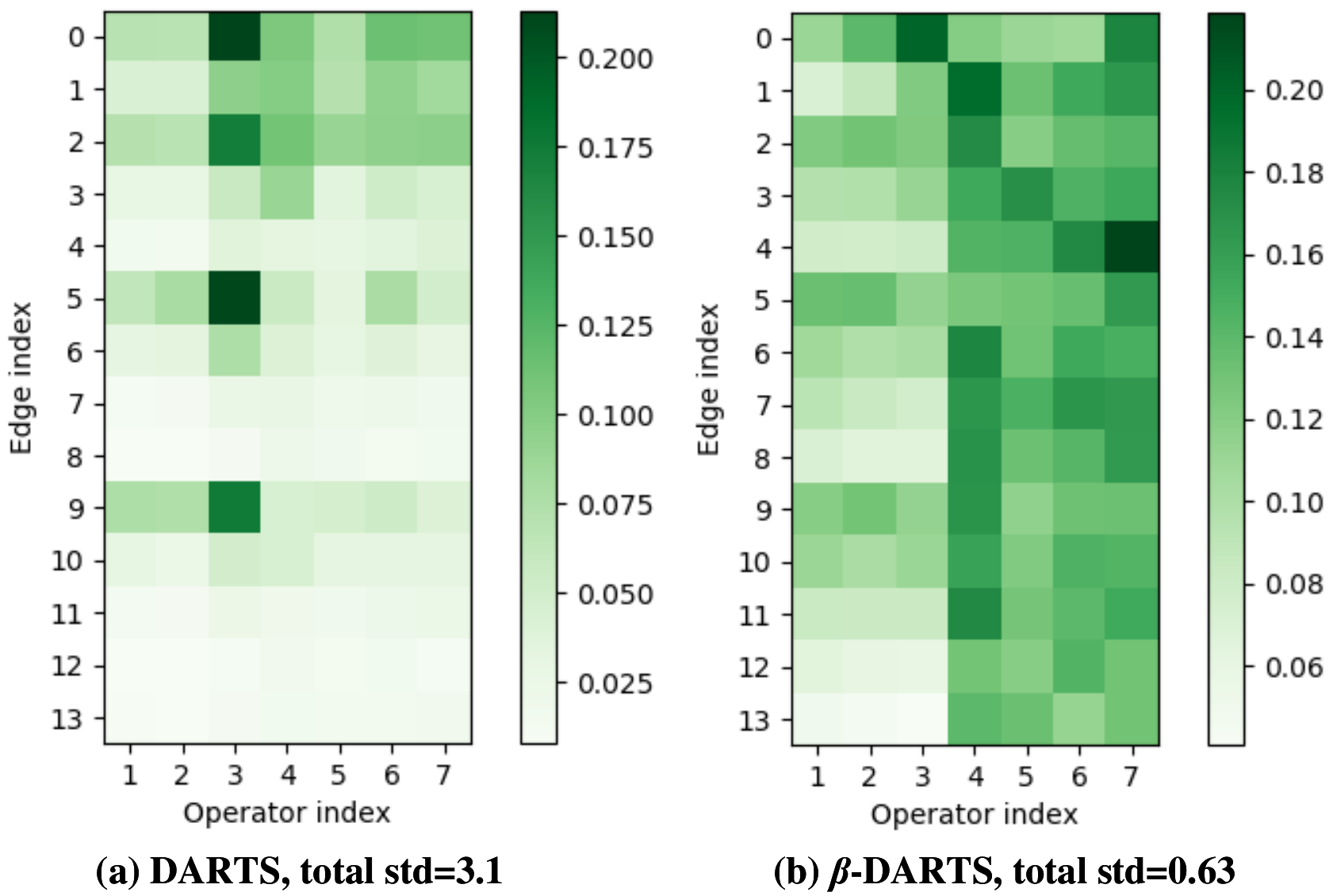}
	\vspace{-8pt}
	\caption{The beta distribution of normal cell learned by DARTS and $\beta$-DARTS, on the original search space in CIFAR-10. The operator indexes 1/2/3 mean the max pool/avg pool/skip connect, while others are the parametric operators. The total std is calculated by the sum of the standard deviation of all edges under the edge independence assumption.} %
	\label{fig:4}
	\vspace{-10pt}	
\end{figure}

\begin{figure*}[t] 
	\centering
	\includegraphics[width=1.0\linewidth]{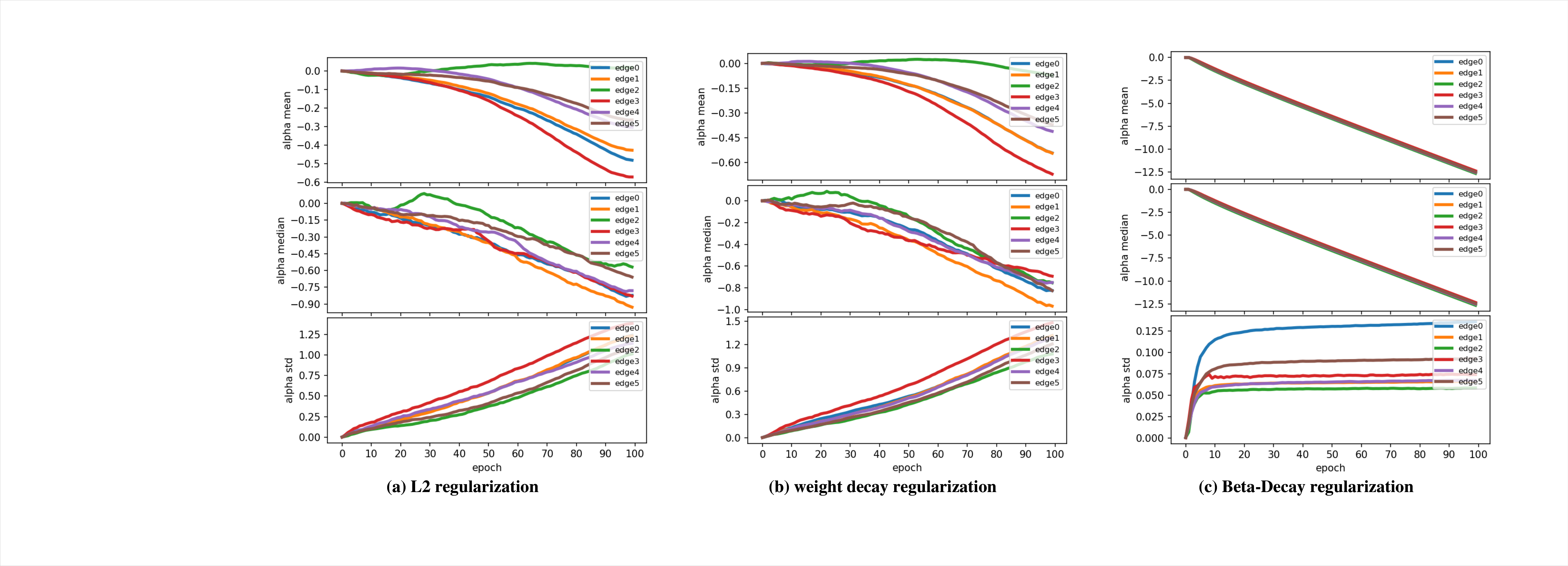}
	\vspace{-12pt}
	\caption{The alpha statistical characteristics (i.e. mean, median and standard deviation) of different edges of each epoch when searching on NAS-Bench-201 benchmark with (a) L2 regularization, (b) weight decay regularization, and (c) Beta-Decay regularization.}
	\label{fig:2}
	\vspace{-10pt}	
\end{figure*}

\noindent\textbf{Stronger Robustness.} According to the theorem revealed by recent work~\cite{prdarts}, the convergence of network weights $w$ can heavily rely on $\beta_{skip}$ in the supernet. In details, supposing that there are three operations (convolution, skip connection and none) in the search space and the training loss is MSE, when fixing architecture parameters to optimize network weights via gradient descent, at one step the training loss can be reduced by ratio  $(1-\eta_w\varphi/4)$ with \bp{a} probability of at least $1-\sigma$,where $\eta_w$ is the corresponding learning rate and will be bounded by $\sigma$, and $\varphi$ obeys
\begin{equation} \label{eqn13}
  \begin{split}
    \varphi \propto \sum_{i=0}^{h-2}\left[\left(\beta_{conv}^{(i, h-1)}\right)^{2} \prod_{t=0}^{i-1}\left(\beta_{skip}^{(t, i)}\right)^{2}\right]
  \end{split}
  \vspace{-6pt}
\end{equation}
where $h$ is the number of supernet layers. From Eq.~(\ref{eqn13}), we can see that $\varphi$ depends more on $\beta_{skip}$ than $\beta_{conv}$, which demonstrates that the supernet weights can converge much faster with large $\beta_{skip}$. However, by imposing Beta-Decay regularization, we can redefine Eq.~(\ref{eqn13}) as follows
\begin{equation} \label{eqn14}
\small
  \begin{split}
    \varphi \propto \sum_{i=0}^{h-2}\left[\left(\theta_{conv}^{(i, h-1)}\beta_{conv}^{(i, h-1)}\right)^{2} \prod_{t=0}^{i-1}\left(\theta_{skip}^{(i, h-1)}\beta_{skip}^{(t, i)}\right)^{2}\right]
  \end{split}
  \vspace{-6pt}
\end{equation}
As mentioned before, $\theta$ becomes smaller when $\beta$ is larger and $\theta$ becomes larger when $\beta$ is smaller, which makes the convergence of network weights rely more on $\beta_{conv}$ and less on $\beta_{skip}$. \textit{From the perspective of convergence theorem}~\cite{prdarts}\textit{, the Beta-Decay regularization constrains the privilege of $\beta_{skip}$ and ensures the fair competition among architecture parameters.} As shown in Fig.~\ref{fig:3}, DARTS with L2 or weight decay regularization suffers from the performance collapse issue, while DARTS with Beta-Decay regularization has a stable search process. As shown in Fig.~\ref{fig:4}, original DARTS is dominated by skip connections while $\beta$-DARTS tends to favor parametric operators.

\noindent\textbf{Stronger Generalization.} Referring to~\cite{neyshabur2017exploring} and~\cite{finlay2018lipschitz}, Lipschitz constraint is commonly used to measure and improve the generalization ability of trained deep model. Specifically, suppose the function fitted by a deep model is $f_{w}\left(x\right)$ where $x$ is the input, when $\left\|x_{1}-x_{2}\right\|$ is very small, a well-trained model should meet the following constraint.
\begin{equation} \label{eqn15}
  \begin{split}
    \left\|f_{w}\left(x_{1}\right)-f_{w}\left(x_{2}\right)\right\| \leq C(w) \cdot\left\|x_{1}-x_{2}\right\|
  \end{split}
  \vspace{-6pt}
\end{equation}
where $C(w)$ is the Lipschitz constant. The smaller the constant is, the trained model will be less sensitive to input disturbances and have better generalization ability. 

Furthermore, we can extend this theory to differentiable architecture search. For simplicity, we consider a single-layer neural network, and multi-layer neural network can be solved through step-wise recursive analysis. Suppose the single-layer network is mixed by the operation set $F(x)=\left(f_{1}(x),f_{2}(x),f_{3}(x)\right)$, with the corresponding architecture parameters $\beta=\left(\beta_{1}, \beta_{2}, \beta_{3}\right)$. According to the Cauchy's inequality, we can get the following inequality.
\begin{equation} \label{eqn16}
\small
  \begin{split}
    &\left\|\beta F^\mathrm{T}\left(x_{1}\right)-\beta F^\mathrm{T}\left(x_{2}\right)\right\| \leq\|\beta \|\left\|F^\mathrm{T}\left(x_{1}\right)-F^\mathrm{T}\left(x_{2}\right)\right\| 
  \end{split}
  \vspace{-6pt}
\end{equation}
where $\|\beta\|=\sqrt{\sum \beta_{i}^{2}}$ can be regarded as Lipschitz constant and $\sum \beta_{i}=1$. \textit{As a result, the smaller the measure $\|\beta\|$ is, the supernet will be less sensitive to the impact of input on the operation set, and the searched architecture will have better generalization ability.} As shown in Fig.~\ref{fig:3}, the model searched by $\beta$-DARTS on CIFAR-10 can well generalize to the CIFAR-100 and ImageNet16 datasets and achieve excellent results. As shown in Fig.~\ref{fig:2} and Fig.~\ref{fig:4}, the architecture parameter distribution learned by $\beta$-DARTS maintains a relative small standard deviation, making sure the generalization ability of the searched model.

\begin{table*}[t]
\begin{center}
\vspace{-12pt}
\caption{Performance comparison on NAS-Bench-201 benchmark~\cite{dong2020bench}. Note that $\beta$-DARTS only searches on CIFAR-10 dataset, but can robustly achieve new SOTA on CIFAR-10, CIFAR-100 and ImageNet16-120. Averaged on 4 independent runs of searching.}
\vspace{-6pt}
\label{tab:1}
\small
\begin{tabular}{cccccccc}
\hline
\multirow{2}{*}{Methods} & \multirow{2}{*}{\begin{tabular}[c]{@{}c@{}}Cost\\ (hours)\end{tabular}} & \multicolumn{2}{c}{CIFAR-10}              & \multicolumn{2}{c}{CIFAR-100}             & \multicolumn{2}{c}{ImageNet16-120}        \\ \cline{3-8} 
                         &                                                                         & valid               & test                & valid               & test                & valid               & test                \\ \hline
DARTS(1st)~\cite{darts}               & 3.2                                                                     & 39.77±0.00          & 54.30±0.00          & 15.03±0.00          & 15.61±0.00          & 16.43±0.00          & 16.32±0.00          \\
DARTS(2nd)~\cite{darts}               & 10.2                                                                    & 39.77±0.00          & 54.30±0.00          & 15.03±0.00          & 15.61±0.00          & 16.43±0.00          & 16.32±0.00          \\
GDAS~\cite{GDAS}                     & 8.7                                                                     & 89.89±0.08          & 93.61±0.09          & 71.34±0.04          & 70.70±0.30          & 41.59±1.33          & 41.71±0.98          \\
SNAS~\cite{snas}                     & -                                                                       & 90.10±1.04          & 92.77±0.83          & 69.69±2.39          & 69.34±1.98          & 42.84±1.79          & 43.16±2.64          \\
DSNAS~\cite{dsnas}                    & -                                                                       & 89.66±0.29          & 93.08±0.13          & 30.87±16.40         & 31.01±16.38         & 40.61±0.09          & 41.07±0.09          \\
PC-DARTS~\cite{pc-darts}                 & -                                                                       & 89.96±0.15          & 93.41±0.30          & 67.12±0.39          & 67.48±0.89          & 40.83±0.08          & 41.31±0.22          \\
iDARTS~\cite{idarts}                   & -                                                                       & 89.86±0.60          & 93.58±0.32          & 70.57±0.24          & 70.83±0.48          & 40.38±0.59          & 40.89±0.68          \\
DARTS-~\cite{darts-}                   & 3.2                                                                     & 91.03±0.44          & 93.80±0.40          & 71.36±1.51          & 71.53±1.51          & 44.87±1.46          & 45.12±0.82          \\
$\beta$-DARTS               & 3.2                                                                     & \textbf{91.55±0.00} & \textbf{94.36±0.00} & \textbf{73.49±0.00} & \textbf{73.51±0.00} & \textbf{46.37±0.00} & \textbf{46.34±0.00} \\
optimal                  & -                                                                       & 91.61               & 94.37               & 73.49               & 73.51               & 46.77               & 47.31               \\ \hline
\end{tabular}
\end{center}
\vspace{-12pt}
\end{table*}

\subsection{Commonly-used Regularization May Not Work} \label{sec:alpha-beta}
When using L2 regularization on $\alpha$, we can obtain its effect on $\beta$ according to Eq.~(\ref{eqn4}) and Eq.~(\ref{eqn9}), defined as
\begin{equation} \label{eqn5}
\footnotesize
  \begin{split}
    \frac{\bar{\beta}_{k}^{t+1}}{\beta_{k}^{t+1}}=\frac{\sum_{k^{\prime}=1}^{\left | \mathcal{O}\right |} \exp \left(\alpha_{k^\prime}^{t+1}\right)}{\sum_{k^{\prime}=1}^{\left | \mathcal{O}\right |} \left[\exp \left(\mathcal{N}(\alpha_{k}^{t})-\mathcal{N}(\alpha_{k^{\prime}}^{t})\right)\right]^{\lambda \eta_{\alpha}} \exp \left(\alpha_{k^\prime}^{t+1}\right)} 
  \end{split}
  \vspace{-6pt}
\end{equation}

Similarly, when using weight decay on $\alpha$, we can obtain its effect on $\beta$ according to Eq.~(\ref{eqn6}) and Eq.~(\ref{eqn9}), as follows
\begin{equation} \label{eqn7}
\footnotesize
  \begin{split}
    \frac{\bar{\beta}_{k}^{t+1}}{\beta_{k}^{t+1}}=\frac{\sum_{k^{\prime}=1}^{\left | \mathcal{O}\right |} \exp \left(\alpha_{k^{\prime}}^{t+1}\right)}{\sum_{k^{\prime}=1}^{\left | \mathcal{O}\right |} \left[\exp \left(\alpha_{k}^{t}-\alpha_{k^{\prime}}^{t}\right)\right]^{\lambda \eta_{\alpha}} \exp \left(\alpha_{k^{\prime}}^{t+1}\right)}
  \end{split}
  \vspace{-6pt}
\end{equation}

From Eq.~(\ref{eqn5}) and Eq.~(\ref{eqn7}), we can find that: (1) When the values in $\alpha$ are all around 0, achieving the purpose of L2 and weight decay regularization, Alpha regularization has little effect on Beta; while when the values in $\alpha$ are all not near 0, it means that both regularization do not work. (2) For L2 and weight decay regularization on $\alpha$, only when the median of $\alpha$ is equal to 0, Alpha regularization has the same and correct effect as Beta regularization. (3) A large variance of $\alpha$ is undesirable, which conflicts with the purpose of L2 and weight decay regularization, and makes the optimization process more sensitive to the hyperparameters $\lambda$ and $\eta_{\alpha}$. In addition, we show the alpha statistical characteristics when searching with different regularization in Fig.~\ref{fig:2}, we can see that for L2 and weight decay regularization: (1) The mean and median of $\alpha$ continue to decrease and gradually move away from 0. (2) The standard deviation of $\alpha$ increases monotonically. These mathematical and experimental results show that L2 or weight decay regularization commonly used in existing gradient-based methods are not identical to Beta regularization, and may not be effective or even counterproductive. As a comparison, with our proposed Beta-Decay regularization: (1) The mean and median of $\alpha$ are basically equal. (2) When the standard deviation of $\alpha$ increases to a certain extent, it will remain unchanged. 

\section{Experiments} \label{sec:Exp}
In this section, we conduct extensive experiments on various search spaces (i.e. NAS-Bench-201, DARTS, NAS-Bench-1Shot1) and datasets (i.e. CIFAR-10, CIFAR-100, ImageNet) to verify the robustness and generalization of $\beta$-DARTS, and we further give some experimental insights about DARTS' dependence on training and data. The overall process of $\beta$-DARTS is summarized in Alg~\ref{alg:Framwork}. 
\begin{algorithm}[htb]
\caption{$\beta$-DARTS}   
\label{alg:Framwork}   
\begin{algorithmic}[1] 
\REQUIRE ~~\\ 
Architecture parameters $\alpha$; Network weights $w$; Number of search epochs $E$; Regularization coefficient adjustment scheme $\lambda_e, e\in \{1,2,…,E\}$.
\STATE Construct a supernet and initialize architecture parameters $\alpha$ and supernet weights $w$
\label{ code:fram:extract }
\STATE For each $e\in \left[ 1, E \right]$ do   
\STATE ~~~Update architecture parameters $\alpha$ by descending \\ 
       ~~~$\nabla_\alpha \mathcal{L}_{val}+ \lambda_e\mathcal{L}_{Beta} $
\STATE ~~~Update network weights w by descending $\nabla_{w} \mathcal{L}_{train}$
\STATE Derive the final architecture based on the learned $\alpha$.
\end{algorithmic} 
\end{algorithm} 
\subsection{Results on NAS-Bench-201 Search Space} 
\noindent\textbf{Settings.} NAS-Bench-201~\cite{dong2020bench} is the most widely used NAS benchmark analyzing various NAS methods. NAS-Bench-201 provides a DARTS-like search space, containing 4 internal nodes with 5 associated operations. The search space consists of 15,625 architectures, with the ground truth performance of CIFAR-10, CIFAR-100 and ImageNet16-120 of each architecture provided. On NAS-Bench-201, the searching settings are kept the same as DARTS on~\cite{dong2020bench}.

\begin{table*}[t]
\begin{center}

\caption{Comparison of SOTA models on CIFAR-10/100 (left) and ImageNet(right). For CIFAR-10/100, results in the top block are obtained by training the best searched model while the bottom block shows the average results of multiple runs of searching. $^\ddagger$ denotes the results of independently searching 3 times on CIFAR-100 and evaluating on both CIFAR-10 and CIFAR-100, while $^\dagger$ denotes the results on CIFAR-10. Because of the difference on classifiers, the network parameters on CIFAR-100 is slightly more than that of CIFAR-10 (about 0.05M). For ImageNet, the top block denotes networks are directly searched on ImageNet (Img.),  the middle block indicates architectures are searched via the idea of Cross Domain (CD.) using CIFAR-10 and part of ImageNet,  models in the bottom block are transferred from the searching results of CIFAR-10 (C10) or CIFAR-100 (C100). $^\ast$ denotes the model is obtained on a different search space.}
\vspace{-6pt}
\label{tab:2}
\resizebox{\linewidth}{!}{
\begin{tabular}{lcccccllccccc}
\cline{1-6} \cline{8-13}
\multirow{2}{*}{Method} & \multirow{2}{*}{\begin{tabular}[c]{@{}c@{}}GPU\\ (Days)\end{tabular}} & \multicolumn{2}{c}{CIFAR-10} & \multicolumn{2}{c}{CIFAR-100} &  & \multirow{2}{*}{Method}             & \multirow{2}{*}{\begin{tabular}[c]{@{}c@{}}GPU\\ (Days)\end{tabular}} & \multirow{2}{*}{\begin{tabular}[c]{@{}c@{}}Params\\ (M)\end{tabular}} & \multirow{2}{*}{\begin{tabular}[c]{@{}c@{}}FLOPs\\ (M)\end{tabular}} & \multirow{2}{*}{\begin{tabular}[c]{@{}c@{}}Top1\\ (\%)\end{tabular}} & \multirow{2}{*}{\begin{tabular}[c]{@{}c@{}}Top5\\ (\%)\end{tabular}} \\ \cline{3-6}
                        &                                                                       & Params(M)    & Acc(\%)       & Params(M)     & Acc(\%)       &  &                                     &                                                                       &                                                                       &                                                                      &                                                                      &                                                                      \\ \cline{1-6} \cline{8-13} 
NASNet-A~\cite{nasnet}                & 2000                                                                  & 3.3          & 97.35         & 3.3           & 83.18         &  & MnasNet-92$^\ast$(Img.)~\cite{mnasnet}                    & 1667                                                                  & 4.4                                                                   & 388                                                                  & 74.8                                                                 & 92.0                                                                 \\
DARTS(1st)~\cite{darts}              & 0.4                                                                   & 3.4          & 97.00±0.14    & 3.4           & 82.46         &  & FairDARTS$^\ast$(Img.)~\cite{fairdarts}                     & 3                                                                     & 4.3                                                                   & 440                                                                  & 75.6                                                                 & 92.6                                                                 \\
DARTS(2nd)~\cite{darts}              & 1                                                                     & 3.3          & 97.24±0.09    & \textbf{-}    & \textbf{-}    &  & PC-DARTS(Img.)~\cite{pc-darts}                      & 3.8                                                                   & 5.3                                                                   & 597                                                                  & 75.8                                                                 & 92.7                                                                 \\
SNAS~\cite{snas}                    & 1.5                                                                   & 2.8          & 97.15±0.02    & 2.8           & 82.45         &  & DOTS(Img.)~\cite{dots}                          & 1.3                                                                   & 5.3                                                                   & 596                                                                  & 76.0                                                                 & 92.8                                                                 \\
GDAS~\cite{GDAS}                    & 0.2                                                                   & 3.4          & 97.07         & 3.4           & 81.62         &  & DARTS-$^\ast$(Img.)~\cite{darts-}                        & 4.5                                                                   & 4.9                                                                   & 467                                                                  & 76.2                                                                 & 93.0                                                                 \\ \cline{8-13} 
P-DARTS~\cite{pdarts}                 & 0.3                                                                   & 3.4          & 97.50         & 3.6           & 82.51         &  & \multicolumn{1}{c}{AdaptNAS-S(CD.)~\cite{adaptNAS}} & 1.8                                                                   & 5.0                                                                   & 552                                                                  & 74.7                                                                 & 92.2                                                                 \\
PC-DARTS~\cite{pc-darts}                & 0.1                                                                   & 3.6          & 97.43±0.07    & 3.6           & 83.10         &  & \multicolumn{1}{c}{AdaptNAS-C(CD.)~\cite{adaptNAS}} & 2.0                                                                   & 5.3                                                                   & 583                                                                  & 75.8                                                                 & 92.6                                                                 \\ \cline{1-6} \cline{8-13} 
P-DARTS~\cite{pdarts}                 & 0.3                                                                   & 3.3±0.21     & 97.19±0.14    & -             & -             &  & AmoebaNet-C(C10)~\cite{amoebanet}                    & 3150                                                                  & 6.4                                                                   & 570                                                                  & 75.7                                                                 & 92.4                                                                 \\
R-DARTS(L2)~\cite{rdarts}             & 1.6                                                                   & -            & 97.05±0.21    & -             & 81.99±0.26    &  & SNAS(C10)~\cite{snas}                           & 1.5                                                                   & 4.3                                                                   & 522                                                                  & 72.7                                                                 & 90.8                                                                 \\
SDARTS-ADV~\cite{sdarts}              & 1.3                                                                   & 3.3          & 97.39±0.02    & -             & -             &  & P-DARTS(C100)~\cite{pdarts}                       & 0.3                                                                   & 5.1                                                                   & 577                                                                  & 75.3                                                                 & 92.5                                                                 \\
DOTS~\cite{dots}                    & 0.3                                                                   & 3.5          & 97.51±0.06    & 4.1           & 83.52±0.13    &  & SDARTS-ADV(C10)~\cite{sdarts}                     & 1.3                                                                   & 5.4                                                                   & 594                                                                  & 74.8                                                                 & 92.2                                                                 \\
DARTS+PT~\cite{darts+pt}                & 0.8                                                                   & 3.0          & 97.39±0.08    & -             & -             &  & DOTS(C10)~\cite{dots}                           & 0.3                                                                   & 5.2                                                                   & 581                                                                  & 75.7                                                                 & 92.6                                                                 \\
DARTS-~\cite{darts-}                  & 0.4                                                                   & 3.5±0.13     & 97.41±0.08    & 3.4           & 82.49±0.25    &  & DARTS+PT(C10)~\cite{darts+pt}                       & 0.8                                                                   & 4.6                                                                   & -                                                                    & 74.5                                                                 & 92.0                                                                 \\
$\beta$-DARTS$^\ddagger$         & 0.4                                                                   & 3.78±0.08    & 97.49±0.07    & 3.83±0.08     & 83.48±0.03    &  & $\beta$-DARTS(C100)                     & 0.4                                                  & 5.4                                                  & 597                                                 & 75.8                                                 & 92.9                                                 \\
$\beta$-DARTS$^\dagger$        & 0.4                                                                    & 3.75±0.15    & 97.47±0.08    & 3.80±0.15     & 83.76±0.22    &  & $\beta$-DARTS(C10)                    & 0.4                                                   & 5.5                                                   & 609                                                   & 76.1                                                                     & 93.0                                                                     \\ \cline{1-6} \cline{8-13} 
\end{tabular}
}
\end{center}
\vspace{-12pt}
\end{table*}

\noindent\textbf{Results.} The comparison results are shown in Table~\ref{tab:1}. We only search on CIFAR-10 and use the found genotype to query the performance of various datasets. For robustness, our 4 runs of searching under different random seeds always find the same optimal solution, which is very close to the optimal performance of NAS-Bench-201. Moreover, as shown in Fig.~\ref{fig:3}, the performance collapse issue is well solved and $\beta$-DARTS has a stable search process. For generalization ability, we can see that the architecture found on CIFAR-10 achieves consistent new SOTA on CIFAR-10, CIFAR-100 and ImageNet. For dependency on training and data, as shown in Fig.~\ref{fig:3}, the search process reaches its optimal point at an early stage (i.e., before 20 epochs), on different datasets. Such results validate that $\beta$-DARTS has the ability to find the optimal architecture rapidly. More interestingly, we find that the search process of different datasets reach the optimal point in different epochs, although they belong to the same run of searching on CIFAR-10. More similar results can be found in \bp{Appendix A.1}.

\subsection{Results on DARTS Search Space} 
\noindent\textbf{Settings.} Common DARTS search space~\cite{darts} is also popular for evaluating NAS methods. The search space consists of normal cell and reduction cell. Each cell has 4 intermediate nodes with 14 edges, and each edge is associated with 8 candidate operations. On DARTS search space, all the search settings are kept the same as DARTS since our method only introduces the simple regularization. For evaluation settings, the evaluation on CIFAR-10/100 follows DARTS~\cite{darts} and the evaluation on ImageNet follows P-DARTS~\cite{pdarts} and PC-DARTS~\cite{pc-darts}. 

\begin{table}[t]
\begin{center}
 \caption{Influence of different weighting schemes on $\beta$-DARTS.} 
 \vspace{-6pt}
\label{tab:3}
\resizebox{\linewidth}{!}{
\begin{tabular}{ccc}
\hline
Weighting Scheme & CIFAR-10 valid          & CIFAR-10 test           \\ \hline
0-15/25/50/100   & 91.21/91.55/91.55/91.55 & 93.83/94.36/94.36/94.36 \\
5/10/15/25     & 84.96/90.59/91.55/90.59        & 88.02/93.31/94.36/93.31      \\
25-15/10/5/0   & 90.59/87.30/73.58/39.77        & 93.31/90.65/76.88/54.30      \\ \hline
\end{tabular}
}
\end{center}
\vspace{-12pt}
\end{table}
  \begin{figure}[t] 
    \vspace{-3pt}	
	\centering
	\includegraphics[width=3.3in]{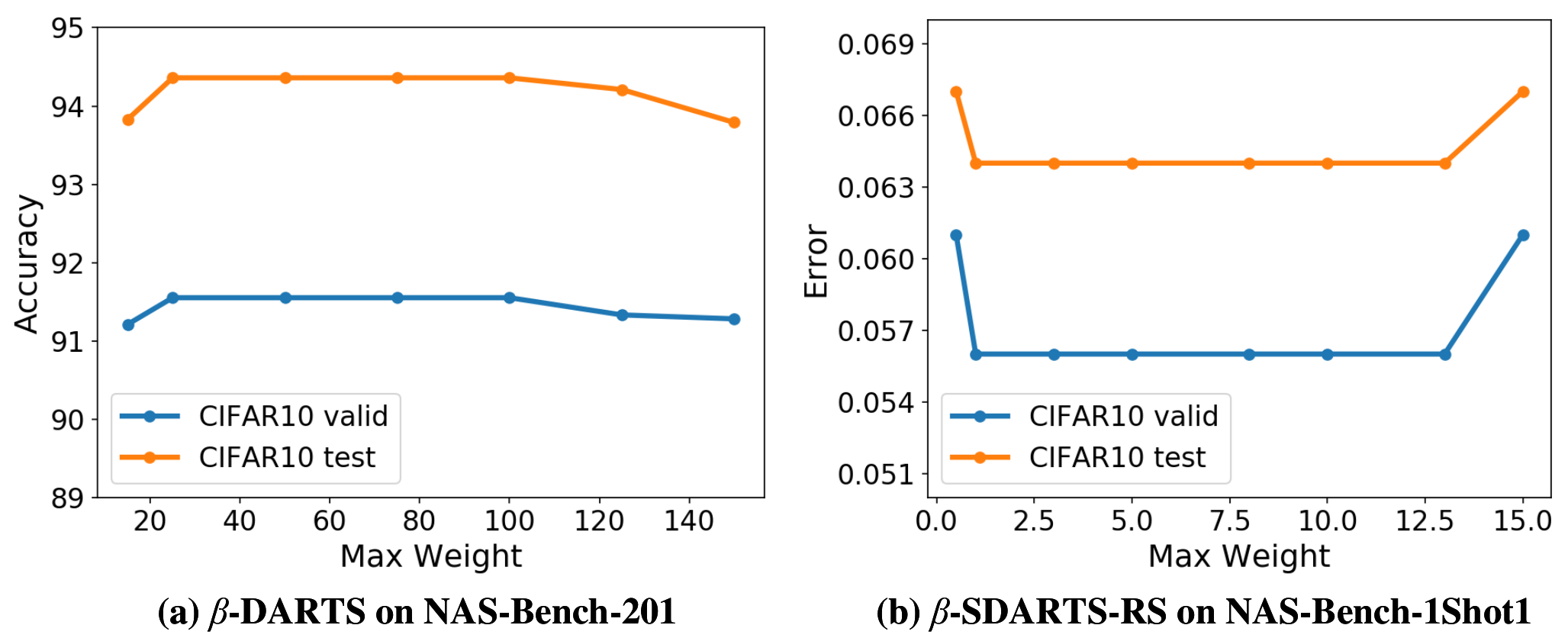}
	\vspace{-22pt}	
	\caption{The effect of different max weight of linear increased weighting schemes on the searching results.}
	\label{fig:5}
	\vspace{-10pt}	
\end{figure}
\begin{table*}[t]
\begin{center}
\caption{The results of different Beta regularization loss with different weighting schemes on NAS-Bench-201 benchmark. Note that we only search on CIFAR-10 dataset, and perform 2 runs of searching under different random seeds.}
\vspace{-6pt}
\label{tab:4}
\small
\begin{tabular}{cccccccc}
\hline
\multirow{2}{*}{Methods} & \multirow{2}{*}{\begin{tabular}[c]{@{}c@{}}Weighting\\ Scheme\end{tabular}} & \multicolumn{2}{c}{CIFAR-10} & \multicolumn{2}{c}{CIFAR-100} & \multicolumn{2}{c}{ImageNet16-120} \\ \cline{3-8} 
                         &                                                                             & valid         & test         & valid         & test          & valid            & test            \\ \hline
DARTS(1st)~\cite{darts}               & 3.2                                                                     & 39.77±0.00    & 54.30±0.00   & 15.03±0.00    & 15.61±0.00    & 16.43±0.00       & 16.32±0.00     \\
Beta-Global                & 0-25                                                                        & 91.55/91.55   & 94.36/94.36  & 73.49/73.49   & 73.51/73.51   & 46.37/46.37      & 46.34/46.34     \\
Beta-Global                & 0-50                                                                        & 91.55/91.55   & 94.36/94.36  & 73.49/73.49   & 73.51/73.51   & 46.37/46.37      & 46.34/46.34     \\
Beta-Global                & 0-75                                                                        & 91.55/91.55   & 94.36/94.36  & 73.49/73.49   & 73.51/73.51   & 46.37/46.37      & 46.34/46.34     \\
Beta-Global                & 0-100                                                                       & 91.21/91.55   & 93.83/94.36  & 71.60/73.49   & 71.88/73.51   & 45.75/46.37      & 44.65/46.34     \\
Beta-Zero                  & 0-25                                                                        & 91.21/90.97   & 93.83/93.91  & 71.60/70.41   & 71.88/70.78   & 45.75/43.77      & 44.65/44.78                 \\
Beta-Zero                  & 0-50                                                                        & 91.55/91.21   & 94.36/93.83  & 73.49/71.60   & 73.51/71.88   & 46.37/45.74      & 46.34/44.65     \\
Beta-Zero                  & 0-75                                                                        & 91.61/91.05   & 94.37/93.66  & 72.75/71.02   & 73.22/71.38   & 45.56/45.23      & 46.71/44.70     \\
Beta-Zero                  & 0-100                                                                       & 91.21/91.21   & 93.83/93.83  & 71.60/71.60   & 71.88/71.88   & 45.75/45.75      & 44.65/44.65     \\ \hline
\end{tabular}
\end{center}
\vspace{-12pt}
\end{table*}
\begin{figure*}[t] 
    \vspace{-3pt}	
	\centering
	\includegraphics[width=1.0\linewidth]{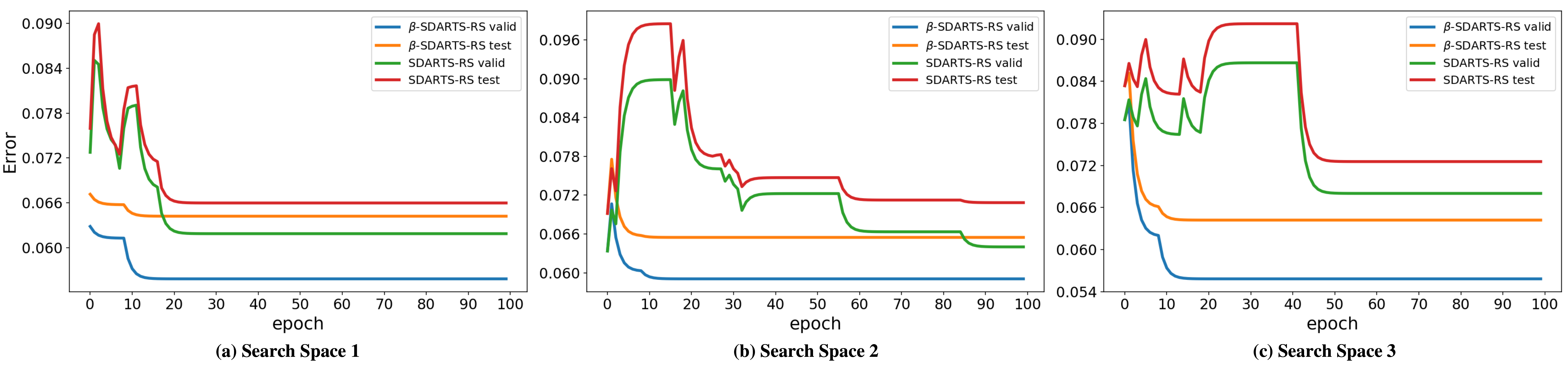}
	\vspace{-12pt}
	\caption{Error of SDARTS-RS and $\beta$-SDARTS-RS on 3 search spaces of NAS-Bench-1Shot1~\cite{1shot1}. The curve is smoothed with 0.5.}
	\label{fig:6}
	\vspace{-10pt}	
\end{figure*}

\noindent\textbf{Results.} The comparison results are shown in Table~\ref{tab:2}. We search on CIFAR-10 or CIFAR-100 while evaluating the inferred architecture on CIFAR-10, CIFAR-100 and ImageNet. For robustness, the average results of multiple independent runs of $\beta$-DARTS achieve the SOTA performance on both CIFAR-10 and CIFAR-100, namely 97.47±0.08\% and 83.48±0.03\%, without extra changes or any cost. For generalization ability, architectures found on CIFAR-100 can still yield a SOTA result of 97.49±0.07\% on CIFAR-10, and models found on CIFAR-10 obtain a new SOTA of 83.76±0.22\%  on CIFAR-100, and networks found on both CIFAR-10 and CIFAR-100 datasets can achieve comparable results on ImageNet with those of directly searching on ImageNet or using cross domain method. 


\subsection{Ablation Study} 
\noindent\textbf{Importance of Increased Weighting Scheme.} We firstly explore the influence of different weighting schemes on $\beta$-DARTS, including linear increased weighting scheme, constant weighting scheme and linear decay weighting scheme. The results are shown in Table~\ref{tab:3}. As we can see, linear decay weighting scheme impedes the effect of regularization, constant weighting scheme is sensitive to the hyperparameter, while linear increased weighting scheme is not only effective but also insensitive to hyperparameter. Besides, combining with the results of Fig.~\ref{fig:3} that the performance on CIFAR-10, CIFAR-100, and Imagenet in single run of searching reach the optimal point in order, we conclude that linear increased regularization coefficient can further maximize the generalization ability of inferred model after the searching performance on current data is maximized, as evidenced by Eq.~(\ref{eqn12}) and Eq.~(\ref{eqn16}).

\noindent\textbf{Wide Range of The Optimal Weight.} We further investigate the optimal max weight of linear increased weighting scheme. The results on CIFAR-10 of NAS-Bench-201 and search space 1 of NAS-Bench-1Shot1 are provided in Fig.~\ref{fig:5}. We can see that the best performance is achieved in a wide range of max weights, namely about 25-100 and 1-13 for $\beta$-DARTS on NAS-Bench-201 and $\beta$-SDARTS-RS on NAS-Bench-1Shot1 respectively. There are similar results on CIFAR-10 and CIFAR-100 in common DARTS search space, as shown in \bp{Appendix A.2}. If not mentioned specially, the default values of max weight for NAS-Bench-201, NAS-Bench-1Shot1, CIFAR-10 and CIFAR-100 are set to 50, 7, 0.5, 5 respectively in all our experiments. Furthermore, comparing Eq.~(\ref{eqn12}) with Eq.~(\ref{eqn5}) and Eq.~(\ref{eqn7}), we find that the normalized values of $\alpha$ in  Eq.~(\ref{eqn12}) has the ability to make sure that the optimization process is not sensitive to the hyperparameter of $\lambda$.

\subsection{Discussions} 
\noindent\textbf{Non-uniqueness.} Actually, the idea of Beta regularization is what really matters, and the way to realize it is non-unique. Here, we show two kinds of variants of Beta regularization loss. Recalling Eq.~(\ref{eqn10}), we can naturally figure out an alternative, using the $\operatorname{smoothmax}$ of all architecture parameters on the entire supernet, namely Beta-Global loss.
\begin{equation} \label{eqn17}
\small
  \begin{split}
    \mathcal{L}_{Beta-Global}&=\operatorname{smoothmax} \left(\begin{array}{r}\alpha_{1}^1,\cdots ,\alpha_{\left | \mathcal{O}\right |}^L\end{array}\right) \\
    &=\log \left(\sum_{l=1}^{L}\sum_{k=1}^{\left | \mathcal{O}\right |}  e^{\alpha_k^l}\right)
  \end{split}
  \vspace{-6pt}
\end{equation}

In addition, by introducing a threshold, we can get the $\operatorname{smoothmax}$ between the threshold and each architecture parameter. We simply set the threshold as 0 in this paper, namely Beta-Zero loss.
\begin{equation} \label{eqn18}
\small
  \begin{split}
    \mathcal{L}_{Beta-Zero}&=\operatorname{smoothmax} \left(\begin{array}{r}0, \alpha_{k}^l\end{array}\right) \\
    &= -\log \left(1+e^{-\alpha_k^l}\right)
  \end{split}
  \vspace{-6pt}
\end{equation}

The results of DARTS with Beta-Global and Beta-Zero regularization loss are shown in Table.~\ref{tab:4}. As we can see, both loss can promote original DARTS by a large margin, while Beta-Global loss that takes the same effect with Beta-Decay loss, can more stably obtain better results than Beta-Zero loss under different weighting schemes. Such results validate that regularizing $\beta$ is important, while the way to achieve it has a lot of room for exploration.


\noindent\textbf{Generality.} Moreover, we utilize NAS-Bench-1Shot1~\cite{1shot1} benchmark and SDARTS-RS~\cite{sdarts} baseline to demonstrate the generality of Beta-Decay regularization. NAS-Bench-1Shot1 contains 3 search spaces, which consist of 6,240, 29,160 and 363,648 architectures with the CIFAR-10 performance separately. On NAS-Bench-1Shot1, both the operator of each edge and the topology of the cell need to be determined. 
We show the search trajectory in Fig.~\ref{fig:6}. On one hand, $\beta$-SDARTS-RS can yield much lower test/validation error than SDARTS-RS across different search spaces. On the other hand, the error of $\beta$-SDARTS-RS keeps decreasing while the error of SDARTS-RS increases first and then decreases, validating the more stable search process of $\beta$-SDARTS-RS. Besides, the search process of $\beta$-SDARTS-RS also reaches its optimal point at an early stage (i.e., around 10 epochs), on different search spaces.

\section{Conclusion} \label{sec:Con}
In this paper, we investigate the explicit regularization on the optimization of architecture parameters of DARTS in depth, which is typically ignored by previous works. Firstly, we identify that L2 or weight decay regularization on alpha commonly used by DARTS and its variants may not be effective or even counterproductive. Then, we propose a novel and generic Beta-Decay regularization loss, for improving DARTS-based methods without extra changes or cost. In addition, we theoretically and experimentally show Beta-Decay regularization can improve both the robustness and the generalization of DARTS. Besides, we find that the proposed search scheme is less dependent on training time and data. Extensive experiments on various search spaces and datasets validate the superiority of our method.

{\small
\bibliographystyle{ieee_fullname}
\bibliography{main_paper}
}

\end{document}